\documentclass{article}

\usepackage{arxiv}

\usepackage[utf8]{inputenc} 
\usepackage[T1]{fontenc}    
\usepackage{hyperref}       
\usepackage{url}            
\usepackage{booktabs}       
\usepackage{amsfonts}       
\usepackage{nicefrac}       
\usepackage{microtype}      
\usepackage{lipsum}		
\usepackage{graphicx}
\usepackage{natbib}
\usepackage{doi}

\usepackage{wrapfig}
\usepackage{amsmath}
\usepackage{multirow}
\usepackage[table,dvipsnames]{xcolor}

\title{Robust EEG-based Emotion Recognition from Coarse Trial-level Labels under Intra-trial Affective Variability}

\author{
    Xiaocong Zeng \\
    School of Mathematics (Zhuhai) \\
    Sun Yat-sen University \\
    Guangdong, 519082, CHINA \\
    \texttt{zengxc3@mail2.sysu.edu.cn} \\
    \And
    Buxuan Wang \\
    School of Computer Science and Engineering \\
    Sun Yat-sen University \\
    Guangdong, 510275, CHINA \\
    \texttt{wangbx29@mail2.sysu.edu.cn} \\
    \And
    Xin Yang \\
    School of Mathematics (Zhuhai) \\
    Sun Yat-sen University \\
    Guangdong, 519082, CHINA \\
    \texttt{yangx367@mail2.sysu.edu.cn} \\
    \And
    Pei Huang \\
    School of Mathematics (Zhuhai) \\
    Sun Yat-sen University \\
    Guangdong, 519082, CHINA \\
    \texttt{huangpei1993@qq.com} \\
    \And
    \href{https://orcid.org/0000-0002-4862-7182}{\includegraphics[scale=0.06]{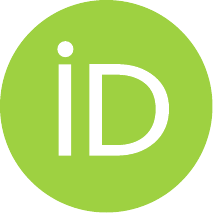}\hspace{1mm}Dongyang Kuang} \\
    School of Mathematics (Zhuhai)\\
	Sun Yat-sen University\\
	Guangdong, 519082, CHINA \\
	\texttt{kuangdy@mail.sysu.edu.cn} \\
}

\date{}

\renewcommand{\shorttitle}{}

\hypersetup{
pdftitle={A template for the arxiv style},
pdfsubject={q-bio.NC, q-bio.QM},
pdfauthor={David S.~Hippocampus, Elias D.~Striatum},
pdfkeywords={First keyword, Second keyword, More},
}

\begin{document}
\maketitle

\begin{abstract}
    EEG-based emotion recognition has become an active research area in brain–computer interfaces. However, fine-grained affective annotation is costly, so most datasets provide only a global label for each long EEG trial. For practical emotion recognition, trials are commonly divided into short temporal windows, with the global label propagated to all windows for training. This practice treats trial-level emotion as temporally uniform, although affective responses may fluctuate within a trial. Consequently, windows that weakly reflect or deviate from the global emotional state may receive unreliable labels, posing a key challenge to robust EEG-based emotion recognition.
To address this challenge, we propose \textbf{VARY}, a variability-aware robust learning framework for EEG-based emotion recognition from coarse trial-level labels. VARY incorporates a Two-stage Sample Selection (TSS) strategy to identify EEG windows with reliable emotional labels, and introduces MixAug, a data augmentation method tailored for EEG to improve model robustness under limited and unreliable supervision.
We comprehensively evaluate VARY under the setting of label-noise learning on two public datasets, using three network architectures and multiple label-noise settings.
Our method demonstrates superior performance, achieving a maximum accuracy gain of 11.54\% on the DREAMER dataset and 5.39\% on the SEED dataset compared with the best-performing baseline, highlighting the effectiveness of VARY in learning from unreliable trial-level labels under intra-trial affective variability.
\end{abstract}


\section{Introduction}
\label{sec:introduction}

Emotion recognition is an important component of affective brain–computer interfaces and human-centered neural engineering systems, enabling machines to infer and respond to users’ affective states \cite{cowie2001emotion,dzedzickis2020human,li2022eeg}. While facial expressions, gestures, and speech are commonly used for emotion analysis, these behavioral cues can be voluntarily masked, suppressed, or affected by external conditions. Physiological signals provide a more direct and objective way to characterize affective responses \cite{shen2022contrastive}. Among them, non-invasive electroencephalography (EEG) is particularly attractive because of its portability, safety, and high temporal resolution, making it suitable for dynamic emotion monitoring. In recent years, deep learning has substantially advanced EEG-based emotion recognition by learning discriminative representations from neural signals \cite{li2022eeg,zhang2020investigation}. 

\begin{figure*}[h]
    \centering
    \includegraphics[width=0.9\textwidth]{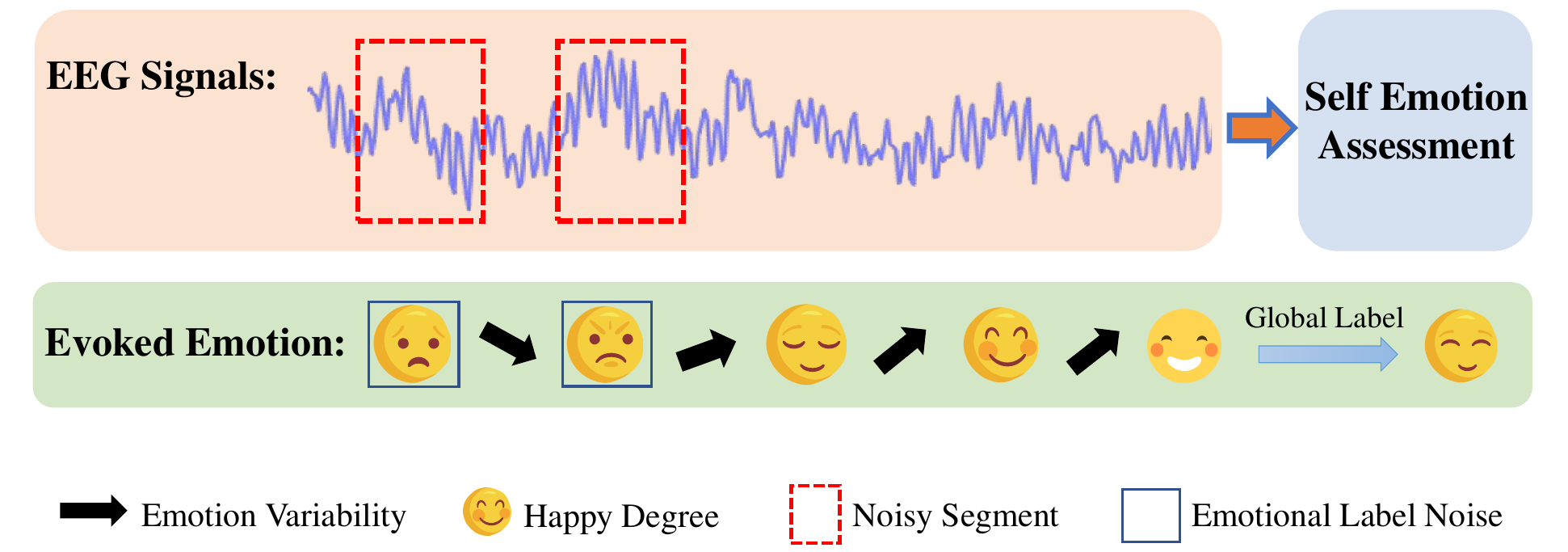}
    \caption{An illustration of window-level label noise under intra-trial affective variability. An EEG trial is assigned a single global label based on post-trial self-assessment, although the evoked emotion may fluctuate over time. When this global label is propagated to all short EEG windows, some windows may receive labels inconsistent with their local affective states. Such local inconsistency can be naturally studied under the setting of label-noise learning.}
    \label{fig:emotion_fluctuation}
\end{figure*}

Despite notable progress, most existing EEG-based emotion recognition datasets still suffer from inherent limitations. Typically, these datasets are collected under a paradigm in which participants watch an emotion-eliciting video for a relatively long duration (usually one to several minutes) while their EEG signals are recorded. Afterward, participants either report their overall emotional state for the entire trial or a pre-defined target emotion is assigned as the coarse trial-level label. Consequently, each long EEG recording is associated with only a single global label, without fine-grained annotations for shorter segments. Moreover, due to the lengthy and labor-intensive nature of data collection, each subject usually contributes only a limited number of trials, leading to severe data scarcity. However, in real-world applications, there is often a demand for short-term emotion prediction (e.g., real-time user experience feedback), and deep learning methods require sufficiently large datasets to train reliable models. To address both issues, prior studies have commonly segmented long EEG recordings into shorter windows and assigned each segment the global label of its parent trial.

Although this practice is reasonable from an engineering perspective, it contradicts neuroscientific and psychological priors, and can introduce inaccurate labels into the dataset.
Studies have shown that human emotions fluctuate over time, reflecting inherent temporal dynamics in affective processes \cite{kuppens2017emotion,waugh2015temporal,puccetti2022temporal}. For example, when watching a video—even if the overall tone is cheerful—you typically don't feel fully engaged from the very beginning. Instead, your emotional response gradually builds up, reaches a peak, and then subsides. This illustrates that the common practice that local EEG segments inherit the coarse trial-level label of the entire recording is often flawed.
As illustrated in Figure~\ref{fig:emotion_fluctuation}, the mismatch between the true local emotional states and the global label assigned by the subject can introduce noisy labels into the dataset, significantly undermining the reliability of the annotations.

Learning with noisy labels can substantially degrade the generalization ability of neural networks, posing a significant challenge for robust training \cite{zhang2021understanding, arpit2017closer}.
To address the adverse effects of noisy labels, robust learning algorithms have been extensively explored in the field of computer vision~\cite{song2022learning}. However, applying these methods to EEG data presents unique challenges. 
Unlike images or text, EEG signals generally exhibit much lower signal-to-noise ratios (SNR) \cite{jiang2024remonet}, making it difficult for models to learn effectively and to reliably distinguish clean samples from noisy ones—a crucial step in sample selection approaches commonly used in label-noise learning \cite{han2018co,li2020dividemix,yuan2025enhancing}. Furthermore, although data augmentation has been a well-established technique in computer vision to alleviate data scarcity and improve generalization, its application to EEG data remains relatively underexplored \cite{rommel2022data}.


To address the aforementioned challenges, we propose \textbf{VARY}, a variability-aware robust learning framework for EEG-based emotion recognition from coarse trial-level labels. The core of VARY consists of two key components: a \textbf{Two-stage Sample Selection (TSS)} method and an EEG-tailored data augmentation strategy, \textbf{MixAug}. Instead of treating all windows within a trial as equally reliable, TSS explicitly accounts for intra-trial affective variability by estimating the reliability of window-level labels propagated from the global trial label. Specifically, we introduce the Label-Prediction Consistency Criterion (LPCC), which is more suitable for low-SNR EEG signals than the commonly used loss-based criterion for identifying EEG windows with reliable labels. Building on LPCC, TSS performs a two-stage selection process to retain informative and reliable windows while reducing the influence of noisy ones. In parallel, MixAug provides a simple yet effective augmentation strategy tailored to EEG data, further improving model robustness under limited and unreliable supervision.

Due to the limitations of current EEG emotion datasets, ground-truth labels for short segments are generally unavailable, posing challenges for rigorous evaluation. Ideally, models should be trained with coarse trial-level labels and evaluated against ground-truth segment-level labels. In their absence, label-noise learning provides a practical surrogate setting, as propagating global trial-level labels to segments with different local affective states may introduce noisy labels. Since previous studies have shown that emotional fluctuations are relatively minor in controlled laboratory settings \cite{zheng2017identifying}, we use two laboratory-collected datasets, DREAMER \cite{katsigiannis2017dreamer} and SEED \cite{zheng2015investigating}, and treat their trial-level labels as reference labels for testing. We then manually introduce different types and levels of synthetic label noise into the training data while keeping the test labels unchanged. This controlled protocol enables systematic evaluation of model robustness under label noise and highlights the need for future datasets with fine-grained segment-level annotations.

The major contribution of this paper can be summarized as
follows:
\begin{itemize}
    \item We address a rarely explored problem in EEG-based emotion recognition: label noise arising from propagating coarse trial-level labels to short EEG windows, which implicitly treats trial-level emotion as temporally uniform and overlooks intra-trial affective variability.
    \item We introduce VARY, a variability-aware robust EEG-based emotion recognition framework that integrates a Two-stage Sample Selection (TSS) strategy, which explicitly accounts for intra-trial affective variability by estimating the reliability of window-level labels propagated from the global trial label, together with MixAug, a simple yet effective data augmentation technique designed to alleviate data scarcity and enhance model robustness.
    \item We comprehensively evaluate our method under the setting of label-noise learning. Extensive experiments across diverse datasets, network architectures, and types of label noise demonstrate the robustness and superior generalization ability of our proposed method under EEG emotion label noise.
\end{itemize}

\section{Related Work}

\subsection{EEG-based Emotion Recognition}

EEG-based emotion recognition (EER) has become a prominent focus in brain-computer interface (BCI) research, driven by advances in deep learning \cite{li2022eeg,zhang2020investigation}. Convolutional neural networks (CNNs) have been widely adopted, with EEGNet \cite{lawhern2018eegnet} demonstrating strong generalization across BCI tasks. Building on neurological insights, RACNN \cite{cui2020eeg} was designed to capture asymmetric brain activity patterns in emotion processing. For modeling temporal dynamics, recurrent architectures like LSTMs have been employed, as seen in BiDANN \cite{li2018novel}, which also addresses domain adaptation challenges. Additionally, graph-based approaches such as DGCNN \cite{song2018eeg} have emerged to model relationships between EEG channels. Most recently, the Transformer architecture, which has revolutionized fields like natural language processing, is being actively explored for EER \cite{liu2022spatial,gong2023eeg}, offering new opportunities for capturing complex spatial and temporal dependencies in EEG data.

Although deep learning methods have significantly advanced EEG-based emotion recognition, most studies overlook the critical challenge of emotion label noise, which arises from directly assigning global trial-level labels to short EEG segments and disregarding the temporal dynamics of emotion.
While recent studies have begun to address this issue \cite{jiang2024remonet,li2022multi,zeng2025commuting}, it remains largely underexplored and leaves considerable room for improvement, including generalization across different network architectures, model performance, and comprehensive evaluation of model reliability. In this paper, we makes efforts to improve upon all of these aspects by designing a network-agnostic, high-performance algorithm and conducting a comprehensive evaluation.

\subsection{Learning with Noisy Labels}

\label{subsection:lnl}

Learning with noisy labels has been widely explored in computer vision \cite{song2022learning}, with robust loss function design and sample selection emerging as two popular approaches. 
The robust loss function paradigm focus on developing noise-tolerant loss functions with theoretical guarantees. \cite{ghosh2017robust} established sufficient conditions for robust loss functions, highlighting the noise robustness of mean absolute error (MAE), though it struggles with complex data. This led to GCE \cite{zhang2018generalized}, unifying MAE and cross-entropy (CE) into a noise-robust framework, and SCE \cite{wang2019symmetric}, which enhances robustness by combining CE with a reverse term. Recent works further enrich this line. $\epsilon$-softmax \cite{wang2024epsilon} relaxes the symmetric constraint via a controllable one-hot approximation, while Student Loss \cite{zhang2024student} embeds a Student-t prior in feature space to automatically down-weight mislabeled samples. While robust loss functions offer strong theoretical guarantees, they often perform well only in simple scenarios.

Sample selection methods exploit the memorization effect of deep neural networks to retain clean samples and discard noisy ones, thereby mitigating the impact of label noise.
Co-teaching \cite{han2018co} trains two networks collaboratively by exchanging selected samples to reduce error accumulation, while SELF \cite{nguyen2019self} improves selection reliability by stabilizing early predictions through exponential moving averaging. DivideMix \cite{li2020dividemix} further combines Co-teaching with semi-supervised learning, treating noisy samples as unlabeled data via MixMatch \cite{berthelot2019mixmatch}.
More recent methods include CS-Isolate \cite{lin2023cs}, which disentangles content and style with iVAE to preserve hard yet reliable samples, and Self-Filtering \cite{wei2022self}, which replaces the small-loss criterion with fluctuation-based selection and adds confidence regularization. SED \cite{sheng2024foster} adaptively separates clean and noisy samples, relabels them with a mean-teacher framework, and re-weights noisy data, whereas ACT \cite{sheng2024enhancing} co-trains a robust model on clean subsets and a non-robust one on full data, identifying clean samples through agreement signals.
Despite their effectiveness, existing label-noise learning methods are not tailored to EEG signals, where low signal quality, limited samples make robust learning more challenging.

\begin{figure*}[t]
    \centering
    \includegraphics[width=0.95\textwidth]{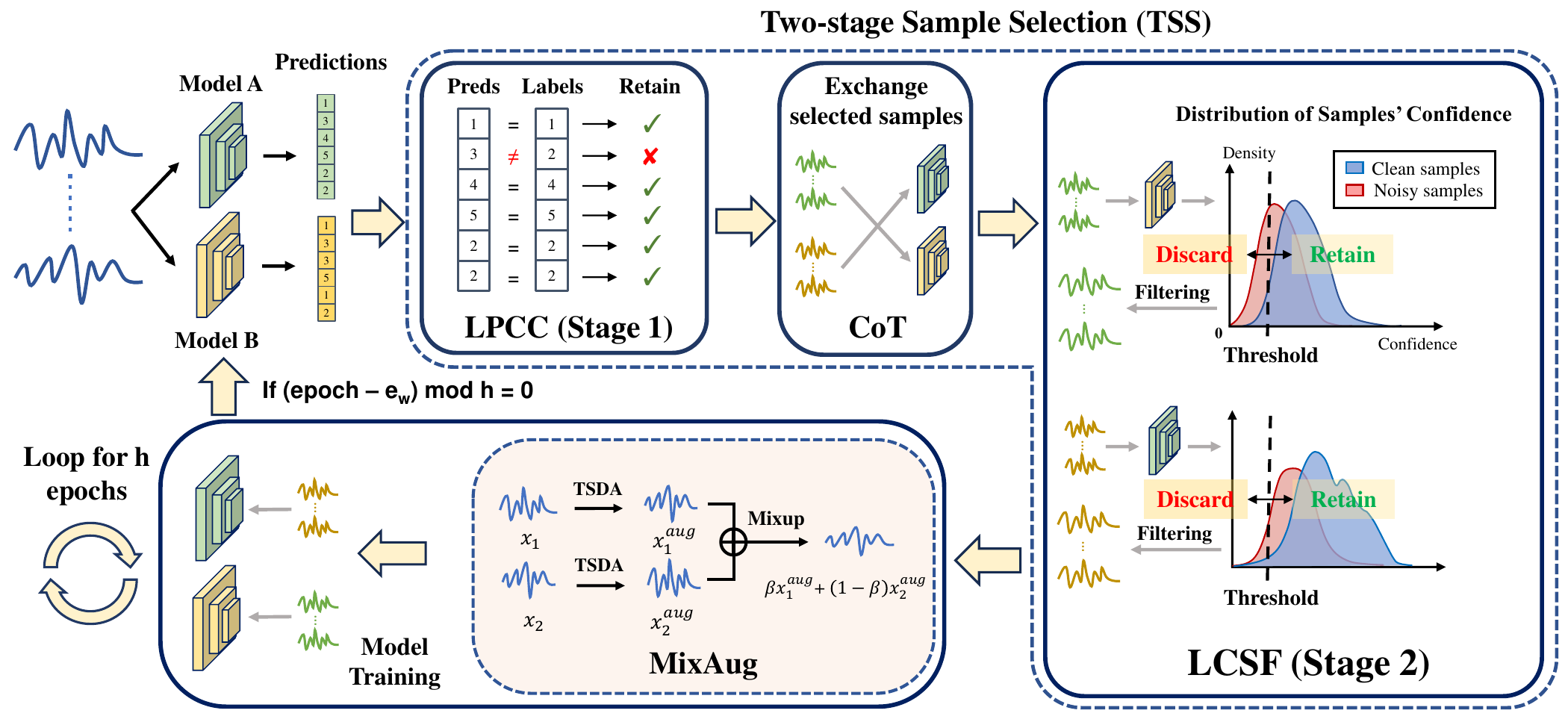}
    \caption{Overview of VARY. We simultaneously train two models (A and B). Following the warm-up phase (Section \ref{warmup}) of length $e_w$ (omitted in this figure), the models employ our proposed Two-stage Sample Selection method (TSS) (Section \ref{method_selection}) to identify clean samples with reliable labels. In the first stage, clean samples are identified based on the Label-Prediction Consistency Criterion (LPCC). In the second stage, the models exchange their selected samples (CoT) and further filter out those with low confidence (LCSF). Once the sample selection is complete, the selected data are augmented using MixAug (Section \ref{mixaug}) (TSDA: time series data augmentation) and subsequently utilized for model training over $h$ epochs. This process repeats iteratively.}
    \label{fig:method}
\end{figure*}

\section{Method}

In this section, we introduce \textbf{VARY}, a variability-aware robust learning framework for EEG-based emotion recognition from coarse trial-level labels.
An overview of \textbf{VARY} is shown in Figure \ref{fig:method}.
We begin by formulating the problem in Section \ref{problem formulation}. Next, we introduce a warm-up stage with a penalty term to guide the model in learning useful information for identifying clean samples while preventing overfitting to noisy labels (Section \ref{warmup}). Follwoing this stage, our method incorporates two key components to achieve robust training: a Two-\textbf{Stage Sample Selection (TSS)} (Section \ref{method_selection}) and \textbf{MixAug} (Section \ref{mixaug}).



\subsection{Problem Formulation}
\label{problem formulation}

We focus on a practical setting of EEG-based emotion recognition where short EEG segments are supervised by coarse trial-level labels, and formulate it as a $K$-class classification problem under noisy labels.
Let the noisy training set be $\mathcal{D} = \{(x_i, y_i')\}_{i=1}^{n}$, where $x_i$ denotes the $i$-th short EEG segment and $y_i'$ is the one-hot label propagated from the global label of its parent trial. Due to the lack of fine-grained affective annotations, the true local label $y_i$ of each segment is unavailable. Under intra-trial affective variability, the propagated label $y_i'$ may be inconsistent with the local affective state reflected in $x_i$, thereby introducing label noise. Our objective is to learn a robust classification model $f(x, \Theta)$ with learnable parameters $\Theta$ that can effectively handle the noisy dataset $\mathcal{D}$, reducing the influence of unreliable segment labels while maintaining strong generalization during testing.
Notably, our method employs a co-training strategy, simultaneously training two networks, $f_1$ and $f_2$. For clarity, we explicitly distinguish between $f_1$ and $f_2$ in the sample selection section (Section \ref{method_selection}). In other sections, however, we use $f$ to denote a generic network when the distinction is not essential.

\subsection{Warm-up with Negative Entropy Penalty}
\label{warmup}
In the initial phase of training, referred to as the warm-up phase, we train the model $f$ on the entire noisy dataset $\mathcal{D}$ to learn informative representations for subsequent sample selection. This phase must be carefully controlled to avoid overfitting to noisy labels. To this end, we encourage the model to maintain higher prediction uncertainty. Entropy is used as a measure of uncertainty, with larger entropy values indicating greater uncertainty. Following \cite{li2020dividemix}, we update the model parameters using the standard cross-entropy loss $\mathcal{L}_c$ augmented with a \textbf{negative entropy penalty (NEP)} $\mathcal{L}_p$, which explicitly promotes higher uncertainty in predictions. The overall warm-up objective $\mathcal{L}_w$ is defined as
\begin{align}
\mathcal{L}_w &= \mathcal{L}_c + \mathcal{L}_p, \\
\mathcal{L}_c &= - \sum_{k=1}^{K} y_i'^k \log f^k(x_i,\Theta), \\
\mathcal{L}_p &= \sum_{k=1}^{K} f^k(x_i,\Theta) \log f^k(x_i,\Theta).
\end{align}
Here, $y_i'^k$ denotes the $k$-th element of the observed one-hot label $y_i'$, and $f^k(x_i,\Theta)$ represents the predicted probability of $x_i$ belonging to class $k$.

\subsection{Two-stage Sample Selection}

\label{method_selection}

To address noisy labels, sample selection has emerged as a widely adopted strategy in computer vision, which aims to retain clean samples while filtering out those with noisy labels \cite{song2022learning}. Prior studies in computer vision have demonstrated that deep neural networks (DNNs) tend to learn clean samples more quickly than noisy ones, leading to lower losses for clean samples \cite{arpit2017closer,zhang2021understanding}. 
Based on this, most sample selection methods adopt the \textbf{Small Loss Criterion (SLC)}, which assumes that samples with lower losses are more likely to be clean.
However, the SLC faces a critical limitation: \textit{it cannot effectively distinguish hard-to-learn clean samples from noisy ones, since both may exhibit relatively high losses} \cite{bai2021me}. This limitation is particularly pronounced in EEG signals, which inherently have low signal-to-noise ratios (SNRs) and are thus difficult to learn. To address this, we propose a novel \textbf{Two-Stage Sample Selection (TSS)} method, grounded in two critical observations that we elaborate on below.

\paragraph{Critical Observation 1 (Prediction Patterns of Hard-to-Learn Samples in the Early Training Stage).}
During the early stages of training, insufficient learning causes hard-to-learn samples (including both noisy samples and intrinsically challenging clean ones) to exhibit two typical patterns: (1) their predictions fluctuate across classes due to incomplete learning, and (2) their predictions may align with the ground-truth label but with low confidence, as the model is influenced by informative signals from other clean samples.
Both patterns leads to relatively high loss values, which makes it challenging to distinguish such samples using the Small Loss Criterion.
Nonetheless, a key distinction exists: while noisy samples under either pattern often exhibit mismatches between labels and predictions, difficult clean samples corresponding to the second pattern consistently preserve label–prediction alignment, albeit with low confidence.

\paragraph{Label-Prediction Consistency Criterion (Stage 1).}
Motivated by \textit{Critical Observation 1}, in the first stage we identify samples whose predictions are consistent with their observed labels as potentially clean samples, a strategy we term the \textit{Label–Prediction Consistency Criterion (LPCC)}.
This criterion helps us effectively separate challenging clean samples from noisy ones, and is formally defined as 
\begin{equation}
    \mathcal{D}_m = \{(x_i, y_i') \in \mathcal{D} \mid \underset{k}{\arg\max} \, y_i'^k = \underset{k}{\arg\max} \; f_m^k(x_i,\Theta)\}, 
\end{equation}
where $m = 1,2$.
Although a similar idea has been explored in \cite{nguyen2019self} within the field of computer vision, it has received limited attention. This is primarily because the high SNR of images results in only a small number of hard-to-learn samples, and the availability of large-scale image datasets makes it possible to discard these samples without notably degrading model performance, thereby allowing SLC-based methods to remain effective. In contrast, our experiments (Section~\ref{ablation_selection}) demonstrate that LPCC not only retains its effectiveness but also surpasses SLC in challenging low-SNR and data-scarce EEG scenarios.

\paragraph{Co-training Strategy.}
\label{cot}
When sample selection relies exclusively on the predictions of a single model, errors can easily accumulate and propagate. To alleviate this problem, we adopt a \textit{Co-Training (CoT) strategy} \cite{blum1998combining,han2018co}, where two models, $f_1$ and $f_2$, are trained in parallel. This strategy provides two major benefits.
First, after the first stage, the models exchange their selected datasets, $\mathcal{D}_1$ and $\mathcal{D}_2$, to update their parameters, allowing each model to leverage the other’s selections and thereby correct their own potential mistakes. Second, in the subsequent stage, we employ Least Confident Sample Filtering (described below), where each model evaluates its samples using the confidence scores predicted by its peer rather than its own. This cross-model evaluation mitigates bias from repeated self-reliance and results in a more reliable and robust sample selection process.

\begin{wrapfigure}{r}{0.5\textwidth}
    \centering
    \includegraphics[width=0.5\textwidth]{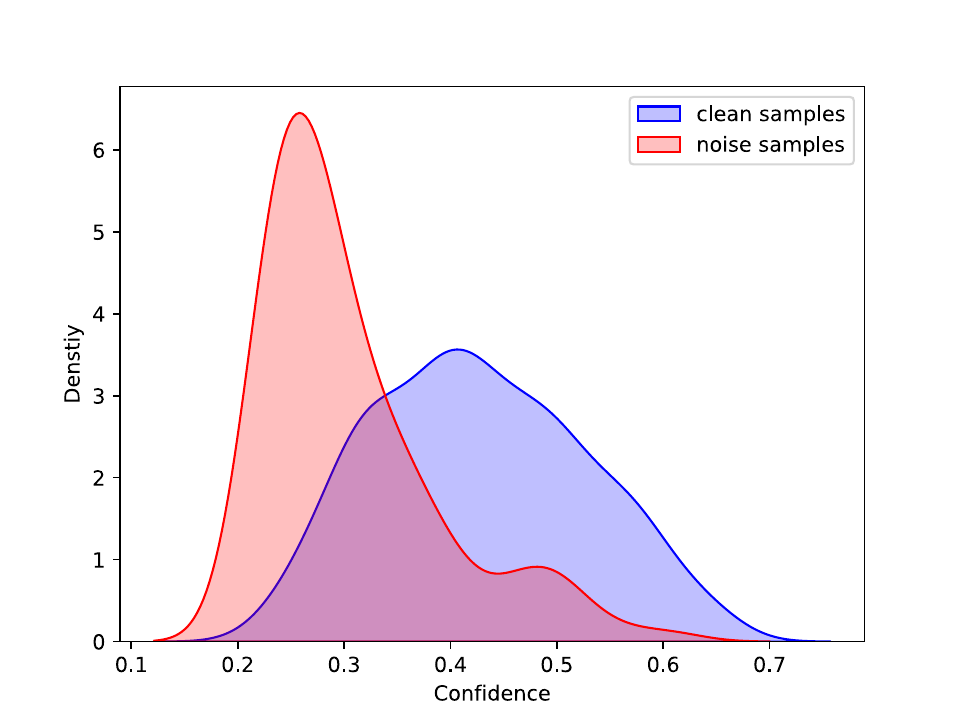}
    \caption{Predicted confidence distributions of clean and noisy samples selected by LPCC in the first stage, from RACNN on the 14th subject of DREAMER with 40\% symmetric noise. Visualized with Seaborn's kdeplot, which applies a Gaussian kernel to smooth histograms.}
    \label{fig:confidence}
\end{wrapfigure}

\paragraph{Critical Observation 2 (Prediction Confidence Distributions of Clean vs. Noisy Samples after LPCC Filtering).}
Due to inevitable model overfitting on noisy labels, some mislabeled samples may persist even after the first selection stage. However, as shown in Figure~\ref{fig:confidence}, these samples generally exhibit lower prediction confidence compared to truly clean ones. Here, confidence is defined as the predicted probability assigned to the \textit{observed label}. For instance, if a sample is annotated as class 5 and the model outputs the prediction $(0, 0.6, 0.1, 0, 0.3)$, its confidence is 0.3.
In our framework, each model relies on the confidence scores predicted by its peer rather than its own (as described in Section~\ref{cot}). Leveraging this insight, we design the Least Confident Sample Filtering procedure to further remove noisy samples.

\paragraph{Least Confident Sample Filtering (Stage 2).}
Building upon Critical Observation 2, we introduce \textit{Least Confident Sample Filtering (LCSF)} to further refine sample selection by discarding the least confident instances. Specifically, we design a gradually relaxed discarding strategy that dynamically adjusts the sample discarding rate according to Equation~\ref{eq_discarding_rate}. In the early stage, stricter thresholds are applied due to the higher noise in the selected data, and they are gradually relaxed as training progresses and the model becomes more reliable. Given the candidate dataset $\mathcal{D}_m$ from the first stage and an initial discarding rate $\alpha$, we sort $\mathcal{D}_m$ in descending order of the confidence scores predicted by the peer model $f_{3-m}$, yielding $\mathcal{D}_m^s$. Then the discarding process is defined as:
\begin{align}
\alpha_e &= \alpha \left( 1 - \sin\left( \frac{\pi}{2} \cdot \frac{e - e_w}{e_k} \right) \right), \label{eq_discarding_rate}\\
\mathcal{D}_m^f &= \{(x_i, y_i') \in \mathcal{D}_m^s \mid i \leq \lfloor (1-\alpha_e) \cdot \lvert \mathcal{D}_m^s \rvert \rfloor \},
\end{align}
where $e_w$ is the length of warm-up phase, $e_k$ the total number of training epochs, and $\alpha_e$ the adaptive discarding rate at epoch $e$.
The final training sets are then constructed as $\mathcal{D}_1^f$ for training $f_2$ and $\mathcal{D}_2^f$ for training $f_1$.

\subsection{Expanding data distribution through MixAug}
\label{mixaug}

Collecting and annotating EEG data is costly, as it demands expensive equipment and lengthy experimental sessions. As a result, EEG datasets often suffer from data scarcity, a problem that becomes even more pronounced after discarding noisy samples in sample selection. While data augmentation (DA) is widely used in computer vision to alleviate this issue, its application to EEG remains relatively unexplored \cite{rommel2022data}. The main challenge lies in expanding the data distribution without compromising the emotional features embedded in EEG signals.

\begin{wrapfigure}{r}{0.5\textwidth}
    \centering
    \includegraphics[width=0.5\textwidth]{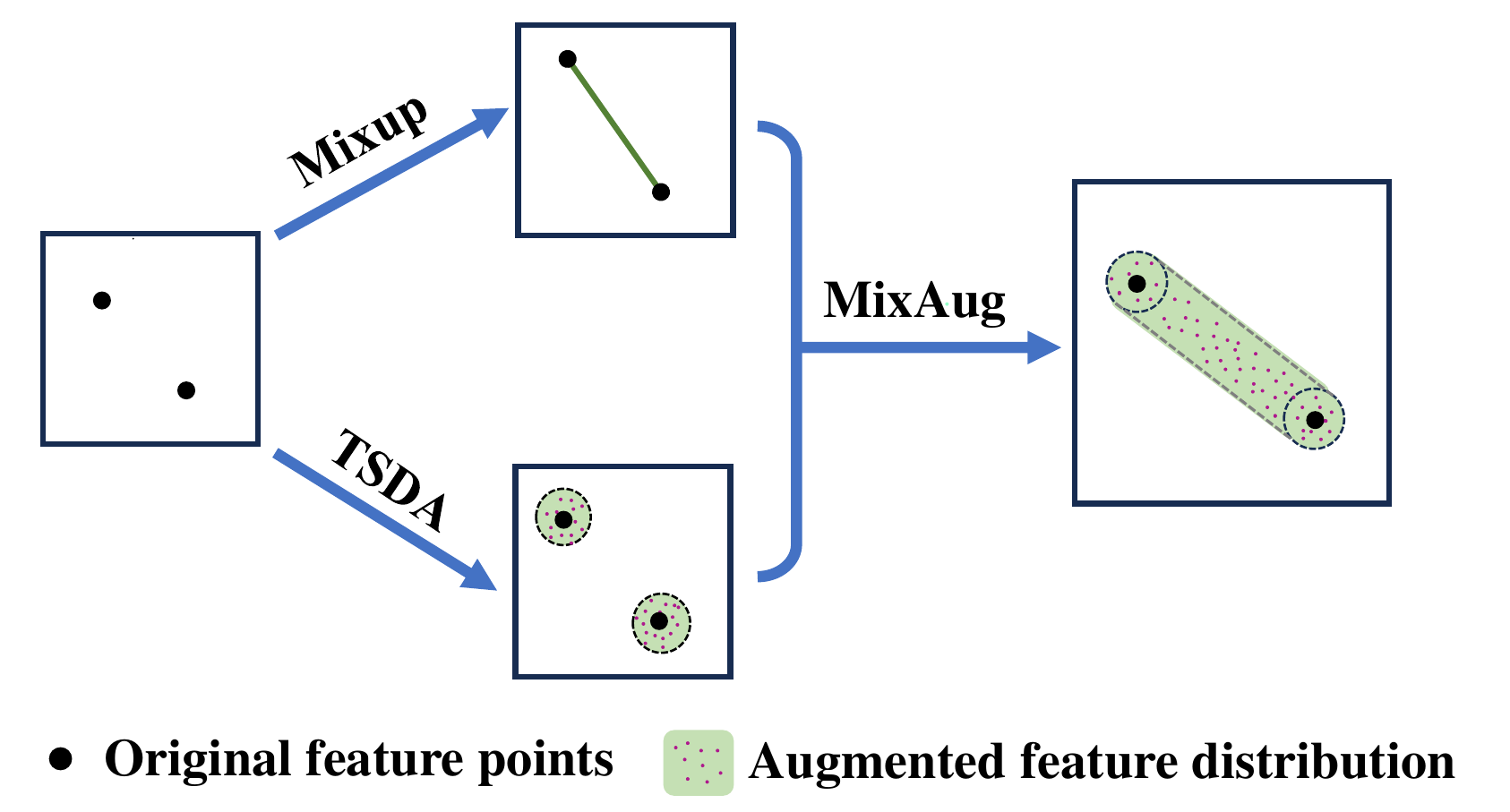}
    \caption{Illustration of MixAug. Mixup expands the data distribution through linear combinations of original samples, while time-series data augmentations (TSDAs) enrich local neighborhoods. By combining both, MixAug further broadens the data distribution.}
    \label{fig:data_distribution}
\end{wrapfigure}

Mixup \cite{zhang2017mixup} is a successful DA technique in computer vision, known for its robustness to label noise. However, our experiments show that its direct application to EEG is less effective (Section \ref{ablation_mixaug}), likely because it does not sufficiently expand the data distribution. Given the temporal nature of EEG signals, as illustrated in Figure \ref{fig:data_distribution}, a natural idea is to combine Mixup with time series data augmentation (TSDA) to achieve a synergistic effect and better expand the data distribution, which we term \textit{MixAug}. Concretely, given two data points $x_1$ and $x_2$, we first apply TSDA to obtain $x_1^{aug}$ and $x_2^{aug}$. Next, we apply Mixup, generating a new sample $(x_m,y_m)$ as:
\begin{align}
x_{m} &= \beta x_1^{aug} + (1 - \beta) x_2^{aug},\quad \beta \sim \text{Beta}(\gamma, \gamma), \\
y_{m} &= \beta y_1' + (1 - \beta) y_2'.
\end{align}

Finally, the training loss for each augmented sample generated by MixAug is computed as:
\begin{equation}
\mathcal{L}_{train} = -\sum_{k=1}^{K} y_{m}^k \log f^k(x_{m}, \Theta).
\end{equation}

However, our experiments in Section~\ref{ablation_mixaug} reveal that the effectiveness of MixAug critically depends on the choice of TSDA. Inappropriate TSDA provides only limited benefits, failing to synergize with Mixup and in some cases even introducing adverse effects. This is likely because unsuitable TSDA cannot adequately preserve the emotional features embedded in EEG signals when combined with Mixup. In contrast, with a well-chosen TSDA, MixAug achieves substantial performance gains, producing a synergistic effect that exceeds the additive contributions of the two methods.

\section{Experiment}

\subsection{Datasets and Backbone Networks}

Emotion quantification can be approached from two perspectives: \textbf{continuous models}, which describe emotions along dimensions such as Valence and Arousal, and \textbf{discrete models}, which categorize emotions into fixed classes. To comprehensively evaluate our method under these two paradigms, we employ two representative datasets corresponding to each type: DREAMER and SEED.

\textbf{DREAMER}~\cite{katsigiannis2017dreamer} is a representative dataset for \textbf{continuous} emotion modeling, consisting of electroencephalogram (EEG) and electrocardiogram (ECG) signals recorded during affect elicitation through audio-visual stimuli. The dataset includes 23 subjects aged between 22 and 23 years old, each of whom was asked to watch 18 movie clips, with each clip lasting between 65 seconds and 393 seconds. After watching each clip, the subjects rated their evoked emotions on the dimensions of Arousal, Valence, and Dominance, using a scale ranging from 1 to 5. The dataset collected 14 channels of EEG signals during the experiment using the 10/20 system, along with 2 channels of ECG signals. The EEG signals recorded at a sampling rate of 128 Hz. Additionally, 60 seconds of baseline signals  were collected before watching each clip.

\textbf{SEED}~\cite{duan2013differential,zheng2015investigating} is a representative dataset for \textbf{discrete} emotion modeling, comprising EEG and eye movement data from 15 subjects (7 males and 8 females). Participants watched 15 emotional movie clips, each approximately 4 minutes long, designed to elicit positive, neutral, and negative emotions. 
The order of clips was arranged to avoid consecutive presentations of the same emotion type.
EEG signals were recorded using a 62-channel ESI NeuroScan System at a sampling rate of 1000 Hz, while eye movement data, including blink, saccade, and fixation information, were captured using SMI eye-tracking glasses.

To evaluate the adaptability of our method across different network architectures, we conduct experiments on three representative types of backbone networks: CNN-based models (RACNN \cite{cui2020eeg} for DREAMER and EEGNet \cite{lawhern2018eegnet} for SEED), a Transformer-based model, and an LSTM-based model.

\begin{table*}[!t]
\caption{Benchmark results of mean test accuracy (\%) for different methods on the \textbf{valence} dimension of the \textbf{DREAMER} dataset. The best results are highlighted in \textbf{bold}, and the second-best results are \underline{underlined}.}
\centering
\begin{tabular}{l |c c c| c c c| c c c }
\toprule
\multirow{2}{*}{Method} & \multicolumn{3}{c|}{RACNN} & \multicolumn{3}{c|}{Transformer} & \multicolumn{3}{c}{LSTM} \\
 & sym 0.2 & sym 0.4 & asym 0.3 & sym 0.2 & sym 0.4 & asym 0.3 & sym 0.2 & sym 0.4 & asym 0.3 \\
\midrule
Regular   & 67.79 & 48.80 & 59.83 & 61.96 & 47.75 & 56.44 & 73.19 & 56.20 & 65.82 \\
SCE       & 66.59 & 45.49 & 55.65 & 70.54 & 56.65 & 63.37 & 79.22 & 71.21 & 72.48 \\
Co-teaching & 81.97 & 70.16 & 75.00 & 79.61 & \underline{73.58} & \underline{74.93} & \underline{88.09} & 75.02 & 78.82 \\
SELF      & 90.54 & 77.21 & 65.63 & \underline{79.64} & 71.15 & 72.18 & 83.70 & 74.92 & 80.20 \\
DivideMix & 90.01 & \underline{86.61} & \underline{86.78} & 76.46 & 69.90 & 72.90 & 87.02 & \underline{82.27} & \underline{85.60} \\
CTW       & 80.13 & 64.31 & 70.96 & 74.73 & 59.11 & 67.78 & 84.80 & 70.11 & 77.33 \\
SED       & \underline{92.09} & 79.31 & 82.69 & 51.90 & 42.33 & 47.03 & 84.47 & 75.91 & 80.81 \\
ACT       & 85.05 & 80.12 & 85.09 & 77.60 & 72.21 & 70.49 & 82.40 & 72.44 & 79.28 \\
\rowcolor{RoyalBlue!20} VARY (ours)     & \textbf{96.48} & \textbf{93.37} & \textbf{94.00} & \textbf{86.69} & \textbf{82.41} & \textbf{81.57} & \textbf{93.18} & \textbf{88.38} & \textbf{90.59} \\
\bottomrule
\end{tabular}
\label{tab:dreamer_valence}
\end{table*}

\begin{table*}[!t]
\caption{Benchmark results of mean test accuracy (\%) for different methods on the \textbf{arousal} dimension of the \textbf{DREAMER} dataset. The best results are highlighted in \textbf{bold}, and the second-best results are \underline{underlined}.}
\centering
\begin{tabular}{l |c c c| c c c| c c c }
\toprule
\multirow{2}{*}{Method} & \multicolumn{3}{c|}{RACNN} & \multicolumn{3}{c|}{Transformer} & \multicolumn{3}{c}{LSTM} \\
 & sym 0.2 & sym 0.4 & asym 0.3 & sym 0.2 & sym 0.4 & asym 0.3 & sym 0.2 & sym 0.4 & asym 0.3 \\
\midrule
Regular     & 67.99 & 49.43 & 59.01 & 63.53 & 47.74 & 57.58 & 73.72 & 56.34 & 65.78 \\
SCE         & 66.18 & 45.19 & 55.30 & 72.56 & 58.19 & 65.34 & 79.89 & 73.67 & 72.85 \\
Co-teaching& 82.15 & 70.99 & 72.40 & 80.40 & \underline{75.32} & \underline{75.33} & \underline{88.01} & 76.43 & 79.25 \\
SELF        & 91.52 & 77.77 & 67.01 & \underline{80.51} & 71.76 & 72.32 & 84.42 & 77.37 & 80.58 \\
DivideMix   & 87.49 & \underline{82.08} & 83.46 & 72.33 & 65.21 & 67.99 & 84.39 & \underline{80.55} & \underline{82.92} \\
CTW         & 80.18 & 63.41 & 69.38 & 74.48 & 58.57 & 67.58 & 83.11 & 68.95 & 74.40 \\
SED         & \underline{92.04} & 80.34 & 81.56 & 53.50 & 43.45 & 48.22 & 85.57 & 78.23 & 81.00 \\
ACT         & 86.10 & 81.85 & \underline{84.67} & 79.33 & 74.09 & 71.72 & 82.79 & 74.32 & 79.49 \\
\rowcolor{RoyalBlue!20} VARY (ours)       & \textbf{96.57} & \textbf{93.62} & \textbf{93.36} & \textbf{86.78} & \textbf{82.77} & \textbf{81.50} & \textbf{93.24} & \textbf{88.76} & \textbf{90.46} \\
\bottomrule
\end{tabular}
\label{tab:dreamer_arousal}
\end{table*}

\subsection{Data Preprocessing and Experiment Settings}

We conduct experiments on the DREAMER~\cite{katsigiannis2017dreamer} and SEED~\cite{duan2013differential,zheng2015investigating} datasets with minimal preprocessing and consistent experimental protocols.

\paragraph{Data Preprocessing.}
For the DREAMER dataset, each session includes EEG signals recorded during baseline (without audio-visual stimuli) and stimuli phases. We normalize the stimuli signals using the mean and standard deviation of the baseline signals, and then scale the data to the range of [-1, 1] by dividing by the maximum absolute value.
For the SEED dataset, which does not include baseline signals, we normalize all data within each session (including both training and test sets) using the mean and standard deviation computed from the training data.
All EEG signals are segmented into non-overlapping 1-second windows, with the model predicting the emotional state for each 1-second segment.

\paragraph{Experiment Settings.}

Emotional perceptions vary significantly across individuals, making subject-independent tasks challenging due to the additional effort required for alignment and evaluation. In contrast, subject-dependent settings provide a more controlled environment, allowing us to focus on the label-noise problem caused by coarse trial-level supervision and intra-trial affective variability. Therefore, all experiments in this study are conducted in a subject-dependent setting.

We conduct a 5-class classification task on DREAMER (valence/arousal scores ranging from 1 to 5) and a 3-class classification task on SEED (positive, negative, neutral). Since ground-truth labels for short EEG segments are unavailable, following common practice, we assign each segment the global label of its parent trial. To obtain relatively reliable reference labels for testing, we use only the last one minute of each trial, which is expected to be relatively cleaner. Specifically, the final 20 seconds are used as the test set, while the preceding 40 seconds serve as the training set. This design is motivated by two considerations: (1) the final segments are temporally closest to participants’ self-reported emotions, and (2) emotional activation typically unfolds gradually over time, making later segments less noisy.

We evaluate the robustness of our method under the label-noise learning setting by introducing two types of artificial noise into the training set: symmetric noise (sym), where labels are randomly flipped with equal probability, and asymmetric noise (asym), where labels are shifted in a circular manner (e.g., 1 $\rightarrow$ 2 $\rightarrow$ 3 $\rightarrow$ 1). This controlled setting serves as a practical surrogate for robustness evaluation when fine-grained segment-level annotations are unavailable, since the studied problem can be viewed as label noise induced by propagating coarse trial-level labels to short segments under intra-trial affective variability.
For fair and reliable comparison, all methods are trained with the same hyperparameters whenever possible. Each experiment is repeated 10 times with different random noise injections, and the reported results are averaged across 10 repetitions and all subjects.

\begin{table*}[!t]
\caption{Benchmark results of mean test accuracy (\%) for different methods on the the \textbf{SEED} dataset. The best results are highlighted in \textbf{bold}, and the second-best results are \underline{underlined}.}
\centering
\begin{tabular}{l |c c c| c c c| c c c }
\toprule
\multirow{2}{*}{Method} & \multicolumn{3}{c|}{EEGNet} & \multicolumn{3}{c|}{Transformer} & \multicolumn{3}{c}{LSTM} \\
 & sym 0.2 & sym 0.4 & asym 0.3 & sym 0.2 & sym 0.4 & asym 0.3 & sym 0.2 & sym 0.4 & asym 0.3 \\
\midrule
Regular     & 66.00 & 44.75 & 55.57 & 51.48 & 42.76 & 47.18 & 63.00 & 44.75 & 54.98 \\
SCE         & 63.99 & 43.21 & 54.12 & 51.00 & 42.41 & 46.81 & 78.50 & 52.37 & 66.09 \\
Co-teaching& 71.60 & 41.88 & 56.04 & 56.62 & 43.24 & 49.95 & 83.29 & 57.08 & 73.37 \\
SELF        & 60.82 & 43.91 & 55.17 & 53.25 & 44.46 & 48.50 & 68.12 & 46.94 & 59.07 \\
DivideMix   & \underline{82.94} & 48.23 & \underline{66.45} & \underline{62.60} & \underline{49.27} & \underline{56.36} & \underline{90.85} & \textbf{62.31} & \textbf{79.23} \\
CTW         & 69.95 & 45.22 & 56.93 & 55.92 & 44.79 & 50.04 & 79.14 & 51.12 & 65.40 \\
SED         & 49.90 & 37.06 & 41.98 & 39.44 & 36.79 & 38.28 & 73.87 & 52.90 & 65.36 \\
ACT         & 73.82 & \underline{49.53} & 62.03 & 51.86 & 43.11 & 47.45 & 70.27 & 49.13 & 61.66 \\
\rowcolor{RoyalBlue!20} VARY (ours)       & \textbf{84.77} & \textbf{54.92} & \textbf{69.66} & \textbf{66.54} & \textbf{50.90} & \textbf{56.53} & \textbf{91.54} & \underline{60.61} & \underline{75.55} \\
\bottomrule
\end{tabular}
\label{tab:seed_acc}
\end{table*}

\subsection{Comparison with State-of-the-Art Methods}

In this section, we evaluate the performance of our proposed VARY against several state-of-the-art methods in label-noise learning.
The baselines include Regular (without any specific technique for label-noise learning), SCE~\cite{wang2019symmetric}, Co-teaching~\cite{han2018co}, SELF~\cite{nguyen2019self}, DivideMix~\cite{li2020dividemix}, CTW~\cite{ma2023ctw}, SED~\cite{sheng2024foster} and ACT~\cite{sheng2024enhancing}.
Most of these baselines were originally developed for computer vision tasks and have demonstrated effective performance on image data, whereas CTW is specifically designed for time-series data.

Table \ref{tab:dreamer_valence} and Table \ref{tab:dreamer_arousal} report the mean test accuracy on the DREAMER dataset for valence and arousal, respectively. 
As shown, VARY consistently achieves the highest accuracy across all settings, with an average improvement of 9.43\% over the second-best baseline. These results highlight the superior performance, robustness, and generalizability of VARY across different emotional dimensions, backbone architectures, and noise. In contrast, competing methods exhibit unstable performance when the backbone or noise condition changes. For example, while SED performs well with RACNN and LSTM-based backbone, its performance significantly degrades with the Transformer-based backbone, suggesting that methods originally developed in the computer vision domain cannot be directly adapted to EEG signals. Although CTW is a time-series based method, it still fails to fully adapt to EEG, highlighting the unique challenges of EEG modeling that VARY is specifically designed to address.

Table \ref{tab:seed_acc} presents results on the SEED dataset, where our method also outperforms baseline methods in most settings. The consistent improvements on both DREAMER and SEED demonstrate that VARY can effectively handle both continuous (DREAMER) and discrete (SEED) emotion modeling.

\begin{figure*}[t]
    \centering
    \includegraphics[width=1.\textwidth]{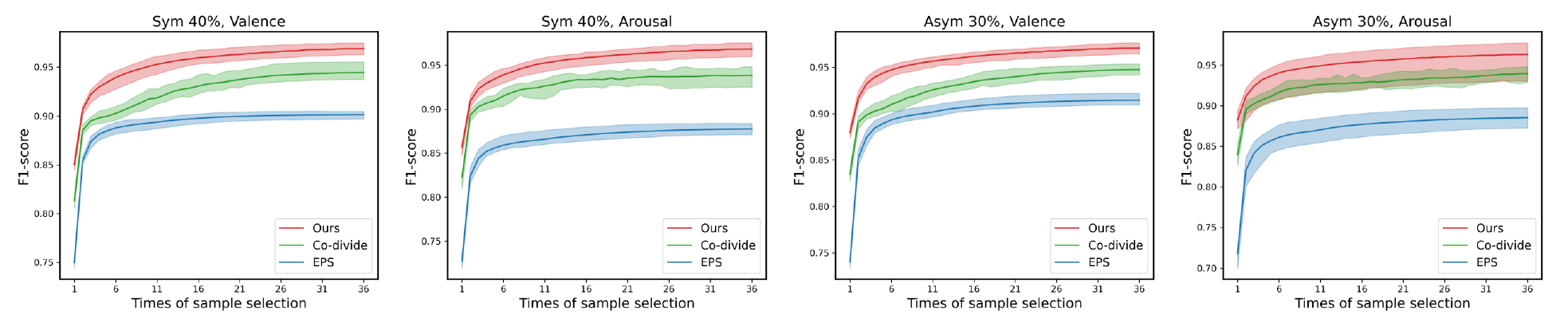}
    \caption{Visualization of the sample selection performance during the training process. The figure reports the F1-scores at each selection step. The F1-score reflects both the cleanliness of the selected samples and the quantity of clean samples chosen, with higher scores indicating better sample selection performance. Our method consistently achieves higher F1-scores compared to the other two SLC-based methods.
    The results are gathered with RACNN and on the DREAMER dataset. The shaded area depicts the range of values observed across multiple runs and subjects.}
    \label{fig:vis_selection}
\end{figure*}

\begin{wraptable}{r}{0.48\textwidth}
\caption{Mean test accuracy (\%) across different sample selection methods on the DREAMER dataset.}
\centering
\resizebox{0.46\textwidth}{!}{
\begin{tabular}{l |c c |c c}
\toprule
\multirow{2}{*}{Method} & \multicolumn{2}{c|}{Valence} & \multicolumn{2}{c}{Arousal} \\
& sym 0.4 & asym 0.3 & sym 0.4 & asym 0.3 \\
\midrule
Co-divide & 89.64 & 91.13 & 86.84 & 87.32 \\
EPS & 86.07 & 87.20 & 84.18 & 84.69 \\
\rowcolor{RoyalBlue!20} TSS (ours) & \textbf{93.37} & \textbf{94.00} & \textbf{93.62} & \textbf{93.36} \\
\bottomrule
\end{tabular}
}
\label{tab:selection}
\end{wraptable}

\subsection{Analysis of Sample Selection Strategy}
\label{ablation_selection}


In Section \ref{method_selection}, we introduced our proposed two-stage sample selection strategy based on the Label–Prediction Consistency Criterion and analyzed why it outperforms methods built upon the Small-Loss Criterion (SLC). In this section, we present experiments to empirically validate its effectiveness.

Concretely, we compare our method with two representative SLC-based approaches. Specifically, we replace the sample selection component (TSS) in our framework with Co-divide \cite{li2020dividemix} and EPS \cite{ma2023ctw}, both of which are carefully designed SLC-based methods, while keeping all other settings unchanged. As shown in Table \ref{tab:selection}, our method consistently outperforms these baselines by a large margin, achieving an average accuracy improvement of nearly 5\% over the second-best method, Co-divide, thereby demonstrating its effectiveness.

To further analyze our approach, we visualize the sample selection results during the training process. In Figure \ref{fig:vis_selection}, we report the F1-score at each sample selection step. The F1-score is defined as:

\begin{align}
\text{F1-score} &= \frac{2 \times \text{CleanRate} \times \text{RecallRate}}{\text{CleanRate} + \text{RecallRate}}\\
\text{CleanRate} &= \frac{\text{number of clean samples selected}}{\text{total selected samples}}\\
\text{RecallRate} &= \frac{\text{number of clean samples selected}}{\text{total clean samples}}
\end{align}

The F1-score accounts for both the purity of the selected samples (CleanRate) and the proportion of clean samples chosen (RecallRate), where higher scores indicate better sample selection performance. Our sample selection method, TSS, consistently achieves higher F1-scores than the other two SLC-based methods. This is because SLC-based methods often struggle to distinguish hard-to-learn clean samples from noisy ones, leading to the selection of many noisy samples and reduced training robustness \cite{bai2021me}. In contrast, our method steadily improves the F1-score, eventually surpassing 95\%. This advantage stems from our two-stage selection strategy: the first stage employs the Label–Prediction Consistency criterion to better separate hard-to-learn clean samples from noise, and the second stage further filters residual noisy samples, ensuring robust performance.

\begin{figure*}[h]
    \centering
    \includegraphics[width=0.95\textwidth]{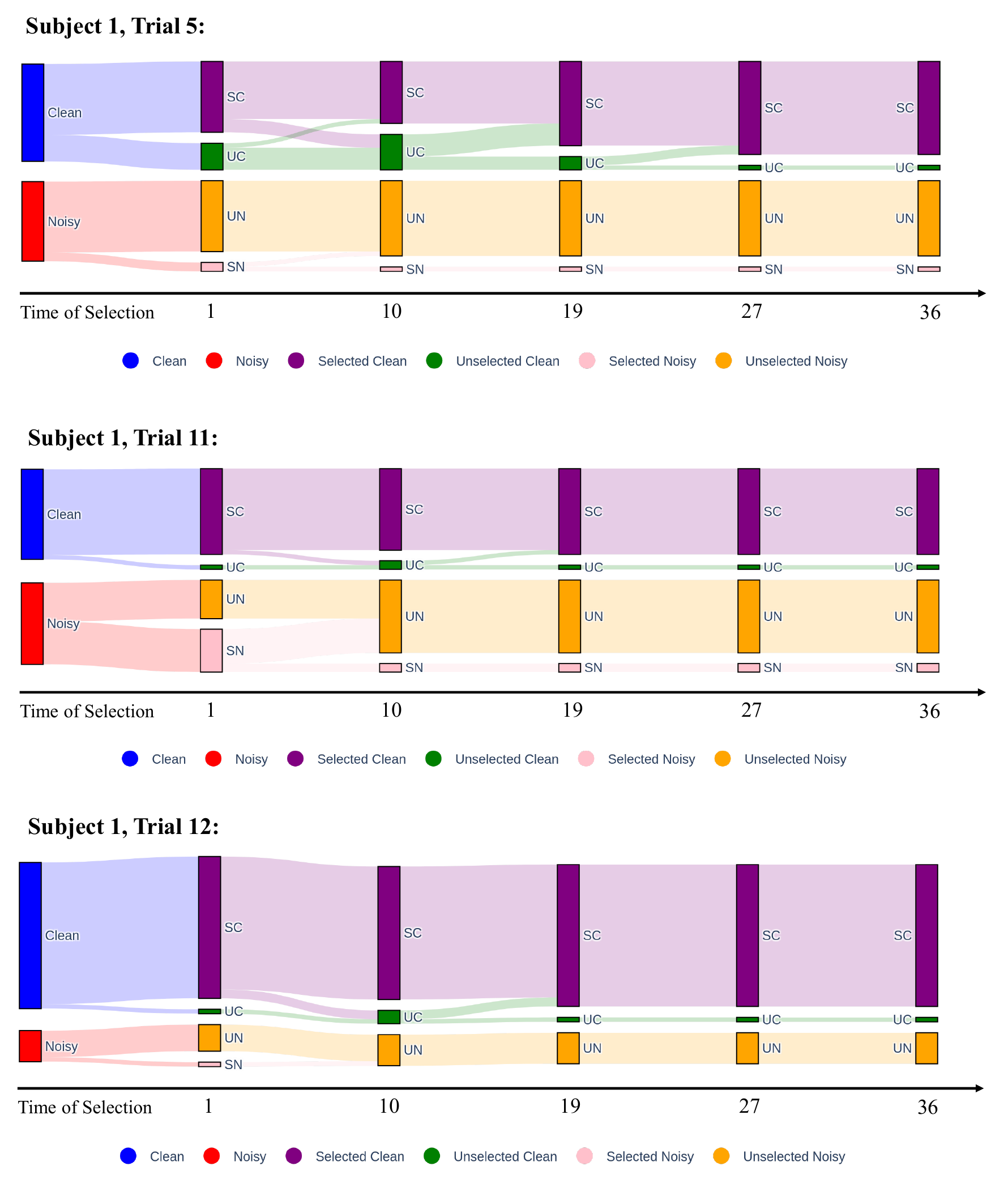}
    \caption{Evolution of sample selection during training. This figure shows the proportions of four sample categories at each selection step—selected clean (correct), unselected noisy (correct), unselected clean (incorrect), and selected noisy (incorrect), as well as their transitions over training. The results demonstrate how the model progressively corrects early selection errors: it initially discards many clean samples and selects noisy ones, but gradually improves its decisions until the selection stabilizes. The results are obtained from Subject 1, trained using RACNN on the DREAMER dataset with 30\% asymmetric noise.}
    \label{fig:vis_selection_1}
\end{figure*}


Furthermore, Figure \ref{fig:vis_selection_1} provides a detailed view of the sample selection process of our method at each step.
This figure presents the proportions and transitions of four sample categories at each selection step: selected clean (correct), unselected noisy (correct), unselected clean (incorrect), and selected noisy (incorrect). The results show that the first selection often performs poorly, with a high proportion of clean samples being incorrectly discarded and many noisy samples being wrongly selected. As training progresses, these errors are gradually corrected, as incorrect samples transition into the correct categories in subsequent selection steps. This demonstrates that our sample selection strategy progressively refines the selection quality, thereby improving the robustness of the model during training.
For instance, in Subject 1, Trial 5, many clean samples are initially unselected, but most of them are correctly retained in subsequent steps. In Subject 1, Trial 11, a large number of noisy samples are initially selected, but they are later identified and removed. Even when correctly identified samples, such as selected clean samples, are mistakenly rejected in a subsequent step, our method can effectively correct these occasional errors in later iterations, demonstrating its robustness to transient misclassifications. Ultimately, the model’s selection decisions stabilize, indicating convergence and reliable performance.

\begin{table*}[t]
\centering
\caption{Time series data augmentation details and settings. (*) indicates that this method is implemented using the Python library tsaug.}
\begin{tabular}{lll}
\toprule
\textbf{Augmentation} & \textbf{Description} & \textbf{Settings} \\ \midrule
AddNoise* & Add random noise to time series. & scale = 0.001 \\
Drift* & \begin{tabular}[t]{@{}l@{}}Drifts the value of time series from its original \\ values randomly and smoothly.\end{tabular} & max\_drift = 0.25, n\_drift\_points = 5 \\
Convolve* & Convolve time series with a kernel window. & window = 'flattop', size = 5 \\
TimeWarp* & Randomly change the speed of timeline. & n\_speed\_change = 5, max\_speed\_ratio = 3 \\
Permutation & \begin{tabular}[t]{@{}l@{}}Randomly shuffles segments of a time series \\ to create new variations.\end{tabular}  & max\_number\_of\_segments = 5\\
\bottomrule
\end{tabular}
\label{tab:augmentation_settings}
\end{table*}


\begin{wraptable}{r}{0.48\textwidth}
\caption{Mean test accuracy (\%) for various data augmentations. Results are gathered with RACNN and on the DREAMER dataset under 40\% symmetric noise. \textbf{w/o: without Mixup, w/: with Mixup.}}
\centering
\resizebox{0.46\textwidth}{!}{
\begin{tabular}{l|cc|cc}
\toprule
\multirow{2}{*}{Method} & \multicolumn{2}{c|}{Valence} & \multicolumn{2}{c}{Arousal} \\
& w/o & w/ & w/o & w/ \\
\midrule
None & \textbf{70.83} & 68.12 & \textbf{70.96} & 69.15 \\
AddNoise &\textbf{71.01} & 69.36 & \textbf{71.44} & 70.27 \\
Drift & \textbf{77.98} & 69.25 & \textbf{77.02} & 69.84 \\
Convolve & 84.66 & \textbf{93.36} & 84.65 & \textbf{93.55} \\
TimeWarp & 79.91 & \textbf{88.34} & 80.33 & \textbf{88.82} \\
Permutation & 84.66 & \textbf{93.37} & 84.65 & \textbf{93.62} \\
\bottomrule
\end{tabular}
}
\label{tab:aug}
\end{wraptable}

\begin{table}[t]
\caption{Mean test accuracy (\%) when each key component of VARY is left-out. Results in \textcolor{red}{red} indicate significant performance drops, highlighting the importance of the corresponding components. Results are gathered with RACNN and on the DREAMER dataset. \textbf{w/o NEP}: without Negative Entropy Penalty in the warm-up phase; \textbf{w/o CoT}: without Co-training strategy; \textbf{w/o LCSF}: without Least Confident Sample Filtering; \textbf{w/o TSS}: without our proposed two-stage sample selection method; \textbf{w/o MixAug}: without MixAug.}
\centering
\begin{tabular}{l |c c |c c}
\toprule
\multirow{2}{*}{Method} & \multicolumn{2}{c|}{Valence} & \multicolumn{2}{c}{Arousal} \\
& sym 0.4 & asym 0.3 & sym 0.4 & asym 0.3 \\
\midrule
\rowcolor{RoyalBlue!20} VARY & \textbf{93.37} & \textbf{94.00} & \textbf{93.62} & \textbf{93.36} \\
w/o NEP & 90.72 & 90.74 & 91.91 & 92.01 \\
w/o CoT & 91.65 & 92.37 & 92.50 & 92.21 \\
w/o LCSF  & 91.74 & 92.81 & 92.09 & 92.49 \\
w/o TSS & \color{red}72.13 & \color{red}77.98 & \color{red}72.79 & \color{red}77.66 \\
w/o MixAug & \color{red}70.83 & \color{red}76.44 & \color{red}70.96 & \color{red}75.42 \\
\bottomrule 
\end{tabular}
\label{tab:ablation-dreamer}
\end{table}

\subsection{Analysis of Data Augmentation Strategy}
\label{ablation_mixaug}


To comprehensively evaluate the effectiveness of MixAug and identify the most suitable time series augmentation techniques for it, we examine five commonly used augmentations: AddNoise, Drift, Convolve, TimeWarp, and Permutation, as detailed in Table~\ref{tab:augmentation_settings}.
The results in Table~\ref{tab:aug} 
reveal several important insights. First, using Mixup alone does not consistently improve performance and can even degrade it, underscoring the importance of the cooperative effect between Mixup and time series augmentation. 
Second, not all augmentation methods synergize well with Mixup. In particular, AddNoise and Drift show limited effectiveness when combined with Mixup, possibly because they disrupt essential emotional features in EEG signals.
In contrast, Convolve, TimeWarp, and Permutation demonstrate significant performance gains, effectively expanding the data distribution while preserving key emotional information, with improvements exceeding 20\% in certain settings. These findings highlight the critical role of carefully designed augmentation strategies in enhancing EEG-based emotion recognition.

\subsection{Component Ablation Studies}
The results presented in Table~\ref{tab:ablation-dreamer} demonstrate that the removal of any key component of our method results in noticeable performance degradation, confirming the importance of each individual module. In particular, omitting the proposed two-stage sample selection (TSS) or the MixAug data augmentation technique leads to the most substantial decline in performance, underscoring their critical roles in
enhancing the robustness and overall effectiveness of our approach. These findings highlight that both TSS and MixAug are indispensable for achieving optimal results, as they respectively ensure high-quality sample selection and effective expansion of the training data distribution.

\section{Conclusion}


In this paper, we focus on a largely overlooked issue in EEG-based emotion recognition: label noise arising from propagating coarse trial-level labels to short temporal windows, which treats trial-level emotion as temporally uniform and overlooks intra-trial affective variability. To address this issue, we propose VARY, a variability-aware robust learning framework for EEG-based emotion recognition from coarse trial-level labels. Our method consists of two key components: a Two-stage Sample Selection (TSS) strategy and a simple yet effective data augmentation technique, MixAug. The TSS strategy addresses the challenge of distinguishing hard-to-learn clean EEG samples from noisy ones by applying a Label-Prediction Consistency criterion in the first stage and confidence-based filtering in the second stage. MixAug effectively expands the data distribution while preserving emotion-related EEG features, thereby mitigating data scarcity and improving robustness. Extensive experiments demonstrate not only the superior performance of VARY under multiple label-noise settings but also the critical contribution of each component. Overall, VARY improves the robustness of EEG-based emotion recognition from coarse trial-level labels and provides a practical step toward reliable affective brain–computer interfaces.

\paragraph{Limitations.}
Our current evaluation is conducted on synthetically generated noisy datasets, which may not fully reflect real-world scenarios. A promising future direction is to collect high-quality EEG datasets with fine-grained, segment-level annotations, enabling more realistic analyses of label noise in practical settings. Moreover, to specifically focus on the label noise problem, we restrict our experiments to the subject-dependent setting. In future work, we aim to extend our approach to the more challenging subject-independent setting, where additional noise may arise from inter-subject variability and subjective biases.

\bibliographystyle{unsrtnat}
\bibliography{references}  






\end{document}